\documentclass[conference]{IEEEtran}
\IEEEoverridecommandlockouts
\usepackage{cite}
\usepackage{amsmath,amssymb,amsfonts}
\usepackage{algorithmic}
\usepackage{graphicx}
\usepackage{textcomp}
\usepackage{xcolor}

\begin{document}

\title{Bio-inspired visual attention for silicon retinas \\ based on spiking neural networks \\ applied to pattern classification
\thanks{This work was supported by the European Union’s ERA-NET CHIST-ERA 2018 research and innovation programme under grant agreement ANR-19-CHR3-0008.
}
}
\author{\IEEEauthorblockN{Amélie Gruel}
\IEEEauthorblockA{\textit{Université Côte d'Azur, CNRS, I3S, France}\\
ORCID 0000-0003-3916-0514\\
amelie.gruel@univ-cotedazur.fr}
\and
\IEEEauthorblockN{Jean Martinet}
\IEEEauthorblockA{\textit{Université Côte d'Azur, CNRS, I3S, France}\\
ORCID 0000-0001-8821-5556\\
jean.martinet@univ-cotedazur.fr}
}


\maketitle

\begin{abstract}
Visual attention can be defined as the behavioral and cognitive process of selectively focusing on a discrete aspect of sensory cues while disregarding other perceivable information. This biological mechanism, more specifically saliency detection, has long been used in multimedia indexing to drive the analysis only on relevant parts of images or videos for further processing.

The recent advent of silicon retinas (or event cameras -- sensors that measure pixel-wise changes in brightness and output asynchronous events accordingly) raises the question of how to adapt attention and saliency to the unconventional type of such sensors' output. 
Silicon retina aims to reproduce the biological retina behaviour. In that respect, they produce punctual events in time that can be construed as neural spikes and interpreted as such by a neural network. 

In particular, Spiking Neural Networks (SNNs) represent an asynchronous type of artificial neural network closer to biology than traditional artificial networks, mainly because they seek to mimic the dynamics of neural membrane and action potentials over time. SNNs receive and process information in the form of spike trains. Therefore, they make for a suitable candidate for the efficient processing and classification of incoming event patterns measured by silicon retinas. 
In this paper, we review the biological background behind the attentional mechanism
, and introduce a case study of event videos classification with SNNs, using a biology-grounded low-level computational attention mechanism, with interesting preliminary results. 
\end{abstract}



\section{Introduction}

Multimedia indexing communities have developed several models and systems dedicated to sound, image, and video, where the visual information is represented by (sequences of) RGB images, optionally including depth. Therefore, traditional video analysis relies on processing full sequences of RGB images.
To limit the amount of data to be processed, a number of approaches inspired by primates' visual attention mechanisms~\cite{itti_computational_2001, martinet_gaze_2009} have proposed to focus the processing only on the most salient parts of the scene. Indeed, the study of many biological organisms highlighted the importance of limiting the processing of sensory regions of interest, thus minimizing the energy spent on this task as well as the reaction delay to what is perceived~\cite{tsotsos_analyzing_1990}. For prey animals for example, this means a quicker detection of a potential predator and a more efficient escape from danger.

Attention was first described as the coalition of ``focalization, concentration and consciousness'' by William James in 1890~\cite{james_principles_1890}. In psychology, this mechanism is now defined as the allocation of limited cognitive processing resources on one or a few relevant environmental elements, while ignoring others~\cite{anderson_cognitive_2005}. It can be either subjective or objective, as well as voluntary or instinctive. 
As technologies for sensory information processing expanded, an increasing number of them assimilated this idea for their own purposes. In the deep learning community for instance, this concept was converted into a neural network's component weighting features by level of importance to a task. Attention can thus be applied to regions in images, words in text, phonemes in speech, etc. For example, some studies aim to exploit this mechanism in order to optimize convolutional neural networks~\cite{vaswani_attention_2017, jetley_learn_2018}. 

In this study, we will focus on visual attention due to its relevance for multimedia indexing and spatio-temporal sensory information processing. 

Silicon retinas represent a new kind of visual sensors, which measures pixel-wise changes in brightness and output asynchronous events accordingly. 
Also known as event-based camera, this novel technology allows for an energy-efficient recording and storage of visuo-spatial data, that is data evolving over time and space.

Traditional methods for standard vision tasks (e.g. recognition, tracking, segmentation, motion analysis, etc.) cannot be applied straightforwardly to event-based cameras due to their unconventional type of output. Silicon retinas bring a large potential in indexing and retrieval, and yet only a limited number of methods have been explored in this direction. In 2020, Gallego et al.~\cite{gallego_event-based_2020} establish {\it inter alia} a state of the art of existing algorithms for feature detection, object tracking, 3D reconstruction and motion segmentation applied to event-data. 

We believe that it would be highly beneficial to leverage the visual attention mechanism in order to optimize the completion of a visual task, such as gesture recognition. 
Via this mechanism, we seek to maximize the resolution of a visual scene, thus increasing the information perceived and processed and obtaining finer results, while minimizing the information bandwidth by decreasing the number of events by bus. 

The following sections are organized as follow. Section II reviews the various organs, cerebral pathways and neural mechanisms involved in
visual attention to paint a global picture of this concept. Section III surveys the related work implementing an attention mechanism using SNNs and/or applied to data output by silicon retinas, as well as a more thorough description of silicon retinas' mode of operation. Finally, section IV proposes several avenues that could be explored to adapt biological attention to computer vision and multimedia indexing, as well as an in-depth account of our ongoing work. 





\section{Biological background}

We believe that the study of the biological aspect of  visual attention will allow for the development of new models applicable to computer vision and multimedia tasks. 

\subsection{Eye and visual cortex}

In this article, we aim to review human visual attention; to this end, we outline what is known about the organs and pathways enabling human vision. 

\textbf{Human eye and retina.} Humans have complex eyes, allowing for colour detection and binocular vision, thus depth perception. It is composed of three layers: the outermost layer is comprised of the cornea and the sclera, whereas the middle layer contains the iris and muscles. The innermost layer, situated at the back of the eye, is mainly composed of the retina which permits the detection of light and colours. The retina contains two types of light-sensing cells: the photo-receptor cells (rods and cones) and the photosensitive ganglion cells. Noticeably, event-based camera aims to emulate this last element's mechanism:
as Steffen et al. described it, the silicon retina is composed of an artificial photoreceptor based on the biological cones, of an adjustable MOS resistor mimicking the retinal horizontal cells and of a bipolar cells-like circuit converting the light signal into ON and OFF events~\cite{steffen_neuromorphic_2019}. 

The inside of the human eye is filled with humour which the light rays cross until they reach the retina. They land mostly on the macula, a spot of the retina responsible for central colour vision. A subdivision of this macula is the fovea, where the cones are more closely packed than anywhere else in the retina. Since cones are responsible for colour distinction, fine detail perception and reaction to image changes, the fovea corresponds to the central spot where the vision is optimal in bright light. Hence any attention mechanism will aim to direct the eyes to point the "foveal gaze" towards the interesting visual feature. It should be noted that the small size of the fovea limits the overall perception of fine details. The organism overcomes this issue by redirecting the eye in saccades, up to 3 times per second. A similar mechanism could be implemented allowing for a higher resolution on the spot targeted by the "foveal gaze", thus letting a larger amount of specifically relevant information through to be processed. 


\textbf{Visual cortex.} Once the visual information has travelled from the eyes and through the thalamus, it is processed by the visual cortex which is situated in the occipital lobe of the brain. Each hemisphere has a visual cortex cross-handling the information output by the opposed eye. The visual data is received firstly by the primary visual cortex (V1); it then journeys through the extrastriate areas V2, V3, V4 and V5.

The frontal eye field (FEF), located in the frontal cortex, has a role in visually guided saccades' target selection and selective spatial attention~\cite{moore_neural_2017}. It encodes spatial information into retinocentric coordinates, that is, coordinates established using the retina as reference. It receives inputs from and projects onto most of the visual cortex. 

According to Knudsen et al.~\cite{knudsen_neural_2018}, high activity in the FEF
brings an increased neuronal response in V4 in a "space-specific, attention-dependant manner". Indeed, the prefrontal cortex integrates the task according to a non-retinocentric frame of reference and translates it into retinocentric signals by sending it to the lateral intra-parietal area. The FEF convert those signals into topographical maps and finally outputs those maps to the retinocentric visual areas of the posterior cortex and superior colliculus.
This may hint at a pathway enabling the influence of saccade-related signals on visual representation. Once understood, this pathway could be implemented as a neural network to specifically direct attention towards the retinocentric space privileged by the FEF.

\textbf{Forebrain and midbrain networks.} Of interest, the forebrain and midbrain networks are strongly involved in attention~\cite{knudsen_neural_2018}. The forebrain selects information based on its relevance to the task at hand or its saliency. It encodes visual information via enhanced distributed representations regulated by heterogeneous dynamics. A similar type of information encoding, with each element interpreted according to its saliency, would speed its processing greatly. 

Regarding the midbrain, it enables spatial attention by monitoring the environment for behaviourally relevant stimuli then directing the gaze to the region of highest interest thanks to a powerful inhibitory competition. It represents the visual input from the retina and the visual forebrain under the form of a topographic map of space following a retinocentric frame of reference. A midbrain-like neural network would allow for the identification of the data relevant to the task at hand in order to optimize the bandwidth of information to process. 

\subsection{Neural mechanisms}

Visual attention can be categorized using three dichotomies, differentiated by distinct and divergent neural mechanisms~\cite{moore_neural_2017}. 
The first dichotomy can be defined between \textit{top-down} and \textit{bottom-up} attention. 
The top-down one corresponds to a selective type of attention depending on a previously set motivation or rule. This "endogenously generated signal" influences visually-driven signals in primates. 
The bottom-up depends on physical saliency, including brightness, movements and colours. 
Those two inter-dependant types of attention interact strongly during visual search, which helps to focus
on salient features relevant to the task at hand and disregard the less fitting ones. 

A second dichotomy exists between \textit{spatial}, \textit{temporal} and \textit{feature-based} attention. Spatial attention is directed towards a specific location in space thanks to the occulomotor system, which leads to the prioritization of an area in the visual field. 
Furthermore, temporal attention focuses on a specific instant in time; this type is often tapped into for video processing and emphasizes critical video frames. It could be particularly interesting to exploit this kind of attention in human action recognition. 
Finally, feature-based attention selects elements based on their resemblance to a behaviorally relevant object, by representing the corresponding features in the prefrontal cortex anterior and the ventral pre-arcuate~\cite{bichot_source_2015}. 

The last dichotomy opposes \textit{overt} and \textit{covert} attention. It respectively corresponds to the presence (overt) or absence (covert) of motor commands leading to saccadic eye movements. Those two types of attention are simultaneous and complementary: the overt attention comes from orientating eye movements, actively guided by salient features determined thanks to concurrent covert attention. %
This is confirmed by the observation of a significant temporal correlation between the visual processing of targets and eye movement~\cite{moore_neural_2017}. 
Interestingly, covert attention is the one most often studied by visual neuroscientists. It improves detection and discrimination of features~\cite{chevallier_covert_2008} both at the fovea and in the visual periphery, thanks to a visual enhancement in V4 and the inferior temporal cortex. 

\subsection{Neuromodulators}

Neuromodulators are a class of extraneuronal molecules released within cortical and subcortical structures which allows and influences the signal transmission between neurons. According to Moore and Zirnsak~\cite{moore_neural_2017}, three neuromodulators have a particularly relevant role in attention: acetylcholine, dopamine and norepinephrine.

Acetylcholine is synthesised by the nucleus basalis of Meynert, the subtantial innominata and the basal forebrain. When collected by a certain type of receptors present on the neuron surface (metabotropic muscarinic or ionotropic nicotinic receptors), it can enhance selective visual attention. Moreover, the release of acetylcholine seems to play a role in the enhanced processing of sensory information: its increase triggers glutamate release from retinal ganglion cells. This release of extracellular molecules by those photosensitive cells boosts their effect and amplifies the visual data captured from the corresponding retinal space. 

Another neuromodulator of interest is dopamine: synthesised by midbrain nuclei, it appears to alter the strength and reliability of converging glutaminergic synapses in the prefrontal cortex, thus affecting the FEF and selective attention. A link has been established between this molecule and Attention Deficit Hyperactivity Disorder (ADHD). 

Also implicated in ADHD, norepinephrine is involved in the selective response to a salient sensory stimuli, depending on the relevance of the stimuli for the task. It is synthesised by the locus cœruleus. 

Much like dopamine pathways were studied and imitated when developing various models of reinforcement learning, it would be highly appealing and relevant to tailor those neuromodulators' mechanisms in order to enhance computational models applied to computer vision tasks. 

\section{Related work}

Applications of visual attention to computer vision and multimedia indexing have been studied for the past two decades. However, the possible impact of spiking neural networks and event-based camera on this domain has been little studied so far. In the following section, we establish a state of the art of attention mechanism models applied to event camera data and/or using spiking neural networks. We also look into previous work applying attention mechanisms to the domain of multimedia. 


\subsection{Attention mechanism using SNNs} 

The use of SNNs instead of traditional artificial neurons have been prioritized in the modelling of attention for some years, maybe due to their common bio-inspiration. Chevallier et al.~\cite{chevallier_efficient_2010} compared the relevance of their use for this task and conclude in favour of spiking neurons over traditional ones, following their previous work on attention in the exploration of a prey-predator environment~\cite{chevallier_distributed_2005}.

SNNs have thus been used to model different types of attention mechanisms, applied to various tasks. For instance, Katayama et al.~\cite{katayama_neural_2004} used SNNs to implement overt attention, through a selective visual attention with gaze shift. This model used two layers: one representing the virtual cortex and the other the hippocampal formation. The correlation of firing times of spikes output by these two layers determines three states: the attention state, the non-attention state and the shift of attention. 

Another application of SNNs is the modelling of covert attention (visual attention without eye movement): Chevallier and Tarroux~\cite{chevallier_covert_2008} use SNNs to extract saliency and focus on the attention of a moving stimulus. This is implemented using a neural filter, a saliency map and a focus map. The filter thresholds enforce a convolution on low and high spatial frequencies. The first map gathers information from visual features on different spatial frequencies thanks to synchrony detectors. The second map finally allows the covert attention using a self-connection mask. In this model, saliencies are temporally coded and arise in hierarchical order, close to the forebrain representation of the visual scene. 

Chick et al.~\cite{chik_selective_2009} defined a model for visual selective attention applied to the sequential selection of objects. This top-down, temporal, feature-based attention uses a 2-layer architecture of inhibitory, excitatory and peripheral neurons. This SNN model used two types of inhibition to implement respectively attention focus and shift thanks to a short term plasticity. 

SNNs can also allow for object detection with saliency indexing, mimicking top-down attention. Such a model has been implemented by Wu et al.~\cite{wu_visual_2013}, in the form of three successive stages: a network inspired by the biological retina extracts the low-level image features, which are then decomposed into multiple visual pathways. Finally, ``\textit{attention area maps}'' similar to the retinocentric output of the FEF are created, approximated by spike rate maps and enabling the detection of regions of interest. 

A model of interest has been defined by Bernert and Yvert~\cite{bernert_attention-based_2018} with an application to spike sorting, a common pattern recognition problem in neurosciences. Spike sorting consists of detecting action potentials emitted by biological neurons, thanks to the classification of the patterns present in the membrane potential measured in said neurons. This unsupervised network reproduces overt temporal attention after a short learning period with little data: an attention neuron modulates intermediate and output layers according to a combined short term plasticity, hysteresis and threshold adaptation mechanism. 

Finally, since curiosity is one of the main drivers of attention, we can mention Shi et al. recent work~\cite{shi_curiosity-based_2020}. To perform object recognition on MNIST data, they regard attention as the novelty of a stimulus. Thus their SNN model learns by estimating the novelty of the visual samples and updating those samples when novelty exceeds a certain threshold. 
According to the authors, the attention mechanism takes place in one of the cerebral pathways responding to curiosity: more specifically, it follows the pathway starting from the ventral tegmental area and reaching the hippocampus via the prefrontal cortex and the precuneus.

As seen above, a few researchers have attempted to implement visual attention in computer vision using SNNs, with interesting results. All types of attention have been studied, on various tasks. However little work has studied the implementation of attention mechanism applied to data produced by silicon retina. The following section describes the corresponding models.   

\subsection{Attention mechanisms applied to silicon retinas}


The idea of a novel bio-inspired event-based sensor, akin to a "\textit{silicon retina}"~\cite{mahowald_silicon_1991}, has been developed since the 1990s. In 2008, Lichtsteiner, Posch and Delbruck presented the first complete design of a Dynamic Vision Sensor (DVS)~\cite{lichtsteiner_128times128_2008} responding only to brightness change in a scene with no consideration for colours, similar to the organic retina. 
An event-based camera~\cite{steffen_neuromorphic_2019} is defined by its capacity to broadcast only relevant information asynchronously, according to the dynamics of the scene. If and only if a pixel perceives a change in luminance above a fixed threshold, an event is emitted with the corresponding polarity.

The main advantages~\cite{gallego_event-based_2020} of such an artificial retina are:
\begin{itemize}
    \item the high temporal resolution, thanks to which an event can be emitted on the timescale of microseconds,
    \item the high dynamic range, thus avoiding motion blur,
    \item the high contrast range, which allows for highly contrasted images avoiding dazzling effect because of sudden illumination changes,
    \item the low latency and asynchronicity enabled by the independence between each pixel,
    \item the absence of redundancy in the information transmitted, as compared to frame-based sensors,
    \item the low power consumption, following the model of biological retinas and substituting the biological photoreceptors by photodiodes in the electrical circuits. 
\end{itemize}


The processing of event-based data often takes place via the adaptation of computer vision models initially dedicated to classic RGB visual information. One example is given by Cannici et al.~\cite{cannici_attention_2018}, who designed two visual attentive models for event-based data. One aims to locate regions of interest using event activity within the field of view, while the other is based on the {\em Deep Recurrent Attentive Writer} (DRAW) neural model. The DRAW model was designed in 2015 by Gregor et al.~\cite{gregor_draw_2015} to generate complex images and emulate the foveation of the human eye, thus implementing a spatial attention mechanism.

Recently, some authors have studied the combination of SNNs used on event-based data in order to convey an attention mechanism. An example of such a combination is given by Bogdan et al.~\cite{bogdan_event-based_2019}, where they develop an unsupervised decomposition of elementary motion detectors. According to them, a "fast localised motion detection is crucial" for an efficient visual attention mechanism. Renner et al. also proposed a combined use of SNNs and silicon retina, but applied to object tracking~\cite{renner_event-based_2019}: they implemented a recurrent SNN emulating the activity of large homogeneous populations of biological neurons. 

\subsection{Attention mechanisms in multimedia}

Itti and Koch presented a complete overview of bottom-up visual attention models in 2001 \cite{itti_computational_2001}. 
They define saliency as a bottom-up type of attention, operating very quickly and independently to the nature of the task. Most of the multimedia domain work on saliency rely upon this definition. 

In 2014, Le Callet and Niebur reviewed the various existing applications of visual attention to multimedia technologies~\cite{le_callet_visual_2013}. Their convincing synthesis of the domain emphasizes the benefit of attention particularly regarding "multimedia delivery, retargeting and quality assessment of image and video, medical imaging, and the field of stereoscopic 3D images applications". It strengthens our conviction that applying attention mechanisms to multimedia, using SNN and silicon retina, is a highly enticing line of scientific inquiry. 

\section{From biological to computational attention}


\subsection{Proposal for biologically-inspired attention models}

The review of attention's biological background presented in Section II allows us to develop some biologically-inspired proposal to adapt the concept of attention to computational models, for purposes of responding to computer vision and multimedia tasks. Indeed, if those tasks are easily achieved by the human brain thanks to attention, the neural mechanisms involved deserve to be studied more thoroughly so as to profit from its efficiency and its robust performance.

Firstly, the forebrain presents a promising implementation of top-down attention. Feature-oriented tasks with specific objectives are common in the domain of computer vision and multimedia (such as gesture recognition, object detection, etc.), and correspond fully to the forebrain's attention use.

An alternate perspective would be to exploit neuromodulator mechanisms. For instance, one could reproduce the acetylcholine's modus operandi to enhance the capture of information from certain regions of interest in visual scenes. Combined with a network mimicking the midbrain and its ability to detect the visual data relevant to the task at hand, this would lead to a higher information processing performance. 

Another important avenue of investigation involves the exploitation of the link between saccadic eye movements and attention, also understandable as the link between overt and covert attention. Such an attention could be achieved by implementing a retinocentric map of saliency, and then processing with a higher resolution on the place where the "\textit{focal gaze}" is directed, i.e. where the saliency is higher. 

\subsection{Adapting an attention model to silicon retinas}

\begin{figure}[b!]
    \centering
    \includegraphics[width=\columnwidth]{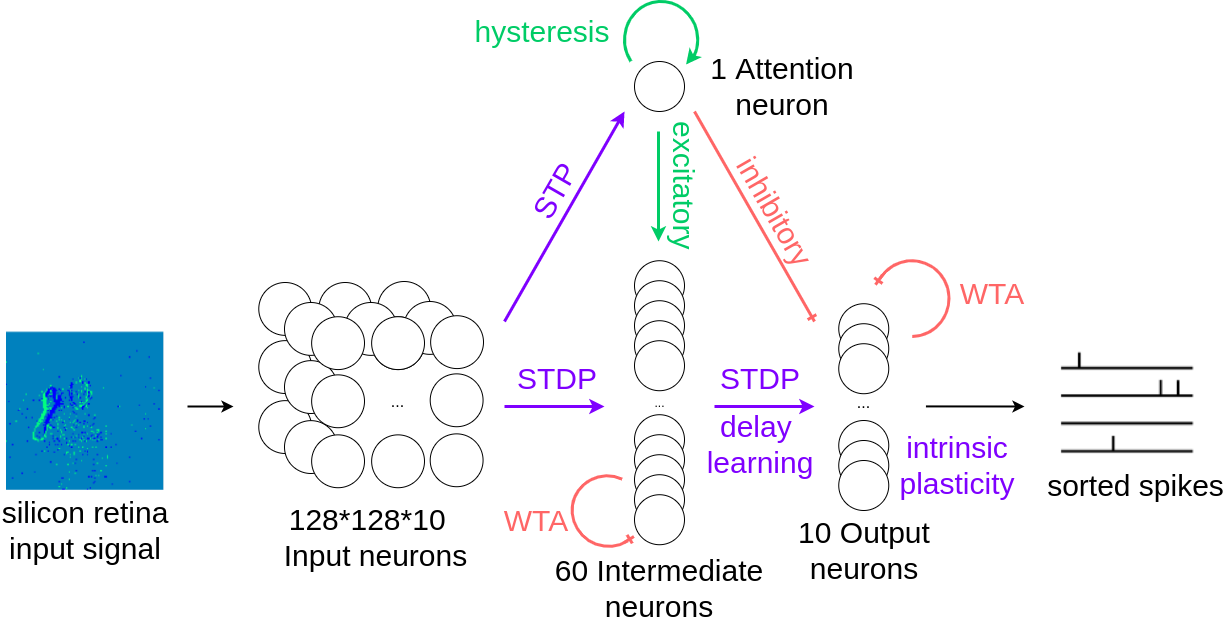}
    \caption{\centering{Overview of the network structure, inspired by the work of Bernert and Yvert~\cite{bernert_attention-based_2018}} and using a sample from the DVS128 Gesture Dataset~\cite{amir_low_2017}}
    \label{struct}
\end{figure}

\begin{figure}[t!]
    \centering
    \includegraphics[width=\columnwidth]{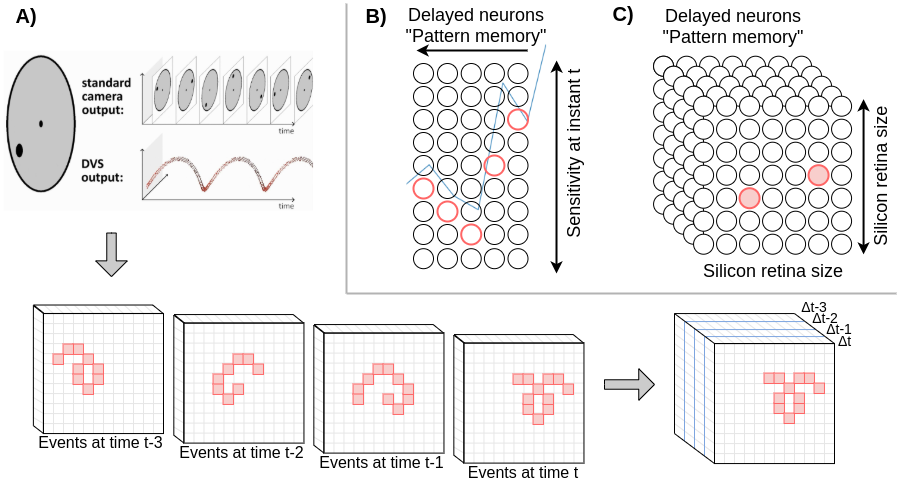}
    \caption{Translating the 2D input implemented by Bernert and Yvert~\cite{bernert_attention-based_2018} (B) into a 3D input adapted to event-based data produced by silicon retinas (C). (A) Illustration of the input for a spiralling pattern, with the concatenation of successive time slices, each conveying events produced during a certain time interval in order to produce the 3D input layer displaying the evolution of events over time. The spiraling events generation figure in (A) has been adapted from Gehrig et al.~\cite{gehrig_daniel_asynchronous_2018}.}
    \label{input}
\end{figure}

This final section presents our ongoing work: we are currently investigating the application of Bernert and Yvert's attention model~\cite{bernert_attention-based_2018} to event data, by translating the original 2-dimensional input into a 3-dimensional one (see Fig.~\ref{input}). We aim to bring out the benefits of using attention for a visual task, thanks to this example of adapting a biologically-inspired attention mechanism to computer vision. 
We intend to use the resulting network as a temporal, bottom-up attentional classifier for event data, such as the DVS128 Gesture Dataset~\cite{amir_low_2017}. 

Bernert and Yvert's model originally implements a classifier fit for a spike-sorting task, aided by an attention neuron supervising the process. Their model receives this 1D signal over time as input, then remits it through two layers via feed-forward synapses implementing specific plasticity rules. 
Each neuron of the Output layer learns a specific pattern and emits a spike when it is detected; the addition of its outputs allows for the detection of an action potential, therefore for spike sorting. 

This classifier is globally supervised by an external 
neuron, modelling the attention. 
The Attention neuron slowly grows accustomed to the input data and in the long term will only react if an input is unexpected. 
It regulates the Intermediate layer in such a way that this last element only processes input data when the Attention neuron is activated; in other words, the classifier only receives input when the Attention neuron deems the data as interesting and relevant. The Attention neuron 
suppresses sporadically the Output neuron: as long as the Attention neuron is active, meaning as long as it receives noteworthy information, the Output layer cannot process the input it is given by the Intermediate layer. This is to ensure the Output layer classifies the entirety of the noteworthy input patterns, not only a sub-pattern. 

\begin{figure*}[h]
    \centering
    \includegraphics[width=\textwidth,]{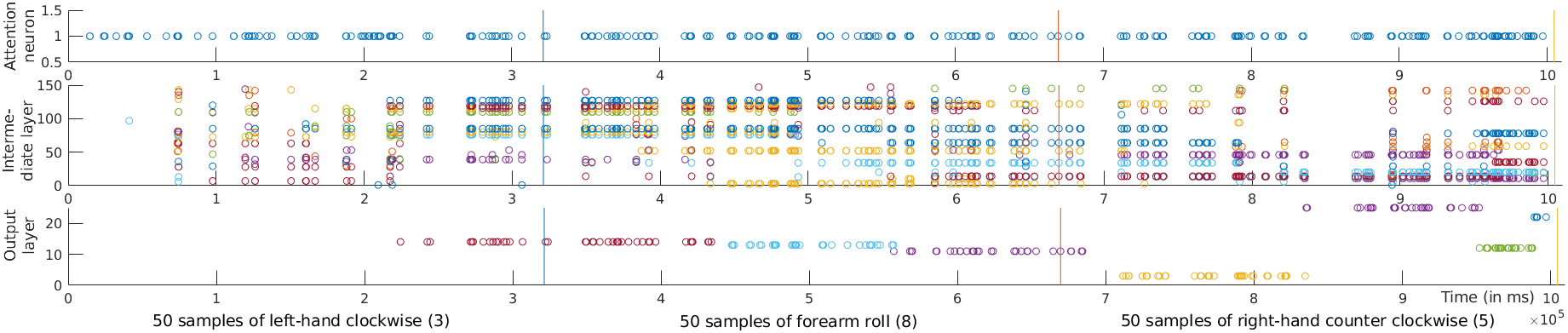}
    \caption{Simulation result on DVS128 Gesture over 1004613 ms, for 50 samples of class 3: \textit{left-hand clockwise}, class 8: \textit{forearm roll} and class 5: \textit{right-hand counter clockwise} (150 in total). Top: Spiking activity of the Attention neuron. Middle: activity of the Intermediate layer. Down: activity of the Output layer.}
    \label{output}
\end{figure*}

In order to process data produced by a silicon retina, the Input layer has been modified starting from Bernert and Yvert's implementation (see Fig.~\ref{input}): one dimension has been added to process visuo-temporal input, meaning input which varies across two dimensions (the length and height of the event-based camera) over time. The original Input layer has a second dimension allowing for a short memory of the input pattern: each column $i$, from right to left, replicates the signal received at time $t_i$ via the activation of \textit{delayed} neurons. This pattern memory is preserved with the transition to 3 dimensions: each layer of the cube will correspond to succeeding time slices, activating according to events received.



Fig.~\ref{output} presents the activation of the 3 layers of our network in response to 150 video samples of three different gestures (50 samples each), originating from the DVS128 Gesture dataset. 
Using a multi-layer perceptron classifier, we can estimate the accuracy of the output of our network under three possible encodings \cite{thorpe_spike-based_2001}. Our current hyperparameters allow for an optimal rate coding accuracy of 0.78 and latency coding accuracy of 0.56 when classifying between 3 categories. The accuracy obtained using rank order coding is only of 0.28. It is interesting to note that the accuracy varies greatly for different sets of hyperparameters and various output's interpretations. 

In future work, we wish to fine-tune the proposed model to obtain more significant results. This will involve fine-tuning the hyperparameters such as the neuron's threshold, the membrane constant and the topology of the network. For example, the Output layer is formed arbitrarily by 10 neurons; this could be optimized for a better fit to our data. We want to investigate how the proposed model can be used in other applications and propose a benchmark of this model compared with other datasets and with other multimedia models of attention. To this end, we need to establish an accuracy measure using SNN or decide on the optimal interpretation.  

\section{Conclusion}
Visual attention selectively focuses on relevant elements of sensory information. Image and video analysis can benefit from the use of attentional mechanisms to improve the processing efficiency, by driving the analysis only on the most pertinent parts of visual scenes.

With the recent availability of bio-inspired silicon retinas, a whole new range of applications emerges. In this paper, we have reviewed the biological background for visual attention, as well as related work regarding SNNs, silicon retinas and multimedia applications.
We put forward some proposals exploiting the known cerebral pathways or adapting existing attention models to data output by silicon retinas. We also introduce a use case of event videos classification using a spiking neural network. The preliminary result achieved strengthens our belief that multimedia analysis can benefit from bio-inspired attentional models.

\bibliographystyle{is-unsrt}
\bibliography{aprovis3d-sophia}

\begin{thebibliography}{10}
\ifx \showCODEN  \undefined \def \showCODEN #1{CODEN #1}  \fi
\ifx \showISBN   \undefined \def \showISBN  #1{ISBN #1}   \fi
\ifx \showISSN   \undefined \def \showISSN  #1{ISSN #1}   \fi
\ifx \showLCCN   \undefined \def \showLCCN  #1{LCCN #1}   \fi
\ifx \showPRICE  \undefined \def \showPRICE #1{#1}        \fi
\ifx \showURL    \undefined \def \showURL {URL }          \fi
\ifx \path       \undefined \input path.sty               \fi
\ifx \ifshowURL \undefined
     \newif \ifshowURL
     \showURLtrue
\fi

\bibitem{itti_computational_2001}
Laurent Itti and Christof Koch.
\newblock Computational modelling of visual attention.
\newblock {\em Nature reviews. Neuroscience}, March 2001.

\bibitem{martinet_gaze_2009}
Jean Martinet, Adel Lablack, Stanislas Lew, and Chabane Djeraba.
\newblock Gaze based quality assessment of visual media understanding.
\newblock {\em IEEE Pacific-Rim Symposium on Image and Video
  Technology-CVIM'09}, 2009.

\bibitem{tsotsos_analyzing_1990}
John~K. Tsotsos.
\newblock Analyzing vision at the complexity level.
\newblock {\em Behavioral and Brain Sciences}, 13\penalty0 (3), 1990.

\bibitem{james_principles_1890}
William James.
\newblock {\em The principles of psychology}.
\newblock H. Holt and Co., 1890.

\bibitem{anderson_cognitive_2005}
John~R Anderson.
\newblock {\em Cognitive {Psychology} and {Its} {Implications}}.
\newblock Worth {Publishers}. 2005.

\bibitem{vaswani_attention_2017}
A.~Vaswani, N.~Shazeer, N.~Parmar, and al.
\newblock Attention {Is} {All} {You} {Need}.
\newblock {\em arXiv}, December 2017.

\bibitem{jetley_learn_2018}
Saumya Jetley, Nicholas~A. Lord, Namhoon Lee, and Philip H.~S. Torr.
\newblock Learn {To} {Pay} {Attention}.
\newblock {\em arXiv}, April 2018.

\bibitem{gallego_event-based_2020}
G.~Gallego, T.~Delbruck, G.~Orchard, and al.
\newblock Event-based {Vision}: {A} {Survey}.
\newblock {\em IEEE PAMI}, 2020.

\bibitem{steffen_neuromorphic_2019}
L.~Steffen, D.~Reichard, J.~Weinland, J.~Kaiser, A.~Roennau, and R.~Dillmann.
\newblock Neuromorphic {Stereo} {Vision}: {A} {Survey} of {Bio}-{Inspired}
  {Sensors} and {Algorithms}.
\newblock {\em Frontiers in Neurorobotics}, 2019.

\bibitem{moore_neural_2017}
Tirin Moore and Marc Zirnsak.
\newblock Neural {Mechanisms} of {Selective} {Visual} {Attention}.
\newblock {\em Annual Review of Psychology}, 68\penalty0 (47-72), January 2017.

\bibitem{knudsen_neural_2018}
Eric~I. Knudsen.
\newblock Neural {Circuits} {That} {Mediate} {Selective} {Attention}: {A}
  {Comparative} {Perspective}.
\newblock {\em Trends in Neurosciences}, November 2018.

\bibitem{bichot_source_2015}
N.P. Bichot, M.T. Heard, E.M. DeGennaro, and R.~Desimone.
\newblock A source for feature based attention in the prefrontal cortex.
\newblock {\em Neuron}, 88, November 2015.

\bibitem{chevallier_covert_2008}
S.~Chevallier and P.~Tarroux.
\newblock Covert {Attention} with a {Spiking} {Neural} {Network}.
\newblock In {\em Intl {Conference} on {Computer} {Vision} {Systems}}, May
  2008.

\bibitem{chevallier_efficient_2010}
S.~Chevallier, N.~Cuperlier, and P.~Gaussier.
\newblock Efficient neural models for visual attention.
\newblock In {\em Computer {Vision} and {Graphics}}, September 2010.

\bibitem{chevallier_distributed_2005}
S.~Chevallier, H.~Paugam-Moisy, and F.~Lemaître.
\newblock Distributed processing for modelling real-time multimodal perception
  in a virtual robot.
\newblock In {\em {PDCN}'2005}, February 2005.

\bibitem{katayama_neural_2004}
K.~Katayama, M.~Yano, and T.~Horiguchi.
\newblock Neural network model of selective visual attention using
  {Hodgkin}-{Huxley} equation.
\newblock {\em Biological Cybernetics}, November 2004.

\bibitem{chik_selective_2009}
D.~Chik, R.~Borisyuk, and Y.~Kazanovich.
\newblock Selective attention model with spiking elements.
\newblock {\em Neural Networks}, September 2009.

\bibitem{wu_visual_2013}
Q.~Wu, T.M. McGinnity, L.~Maguire, R.~Cai, and M.~Chen.
\newblock A visual attention model based on hierarchical spiking neural
  networks.
\newblock {\em Neurocomputing}, September 2013.

\bibitem{bernert_attention-based_2018}
M.~Bernert and B.~Yvert.
\newblock An {Attention}-{Based} {Spiking} {Neural} {Network} for
  {Unsupervised} {Spike}-{Sorting}.
\newblock {\em Intl. J. of Neural Systems}, 2018.

\bibitem{shi_curiosity-based_2020}
M.~Shi, T.~Zhang, and Y.~Zeng.
\newblock A {Curiosity}-{Based} {Learning} {Method} for {Spiking} {Neural}
  {Networks}.
\newblock {\em Front. in Comp. Neuroscience}, 2020.

\bibitem{mahowald_silicon_1991}
M.A. Mahowald and C.~Mead.
\newblock The {Silicon} {Retina}.
\newblock {\em Sci. American}, 1991.

\bibitem{lichtsteiner_128times128_2008}
Patrick Lichtsteiner, Christoph Posch, and Tobi Delbruck.
\newblock A 128x128 120 {dB} 15 us {Latency} {Asynchronous} {Temporal}
  {Contrast} {Vision} {Sensor}.
\newblock {\em IEEE Journal of Solid-State Circuits}, 2008.

\bibitem{cannici_attention_2018}
M.~Cannici, M.~Ciccone, A.~Romanoni, and M.~Matteucci.
\newblock Attention {Mechanisms} for {Object} {Recognition} with
  {Event}-{Based} {Cameras}.
\newblock {\em arXiv}, November 2018.

\bibitem{gregor_draw_2015}
K.~Gregor, I.~Danihelka, A.~Graves, D.J. Rezende, and D.~Wierstra.
\newblock {DRAW}: {A} {Recurrent} {Neural} {Network} {For} {Image}
  {Generation}.
\newblock May 2015.

\bibitem{bogdan_event-based_2019}
P.~Bogdan, G.~García, S.~Davidson, M.~Hopkins, R.~James, and S.~Furber.
\newblock Event-based computation: {Unsupervised} elementary motion
  decomposition.
\newblock In {\em Emerging Technology Conference}, 2019.

\bibitem{renner_event-based_2019}
Alpha Renner, Matthew Evanusa, and Yulia Sandamirskaya.
\newblock Event-{Based} {Attention} and {Tracking} on {Neuromorphic}
  {Hardware}.
\newblock July 2019.

\bibitem{le_callet_visual_2013}
Patrick Le~Callet and Ernst Niebur.
\newblock Visual {Attention} and {Applications} in {Multimedia} {Technologies}.
\newblock {\em Proceedings of the IEEE.}, September 2013.

\bibitem{amir_low_2017}
A.~Amir, B.~Taba, D.~Berg, and al.
\newblock A {Low} {Power}, {Fully} {Event}-{Based} {Gesture} {Recognition}
  {System}.
\newblock In {\em CVPR}. Honolulu, HI, July 2017.

\bibitem{gehrig_daniel_asynchronous_2018}
{Gehrig, D.}, Rebecq H., Gallego G., and {Scaramuzza, D.}
\newblock Asynchronous, {Photometric} {Feature} {Tracking} using {Events} and
  {Frames}.
\newblock {\em ECCV}, 2018.

\bibitem{thorpe_spike-based_2001}
S.~Thorpe, A.~Delorme, and R.~Van~Rullen.
\newblock Spike-based strategies for rapid processing.
\newblock {\em Neural Networks: The Official Journal of the International
  Neural Network Society}, 14\penalty0 (6-7):\penalty0 715--725, September
  2001.
\newblock \showISSN{0893-6080}.

\end{thebibliography}

\end{document}